\pdfoutput=1

\documentclass[11pt]{article}

\usepackage{acl}

\usepackage{times}
\usepackage{latexsym}
\usepackage{amsmath}
\usepackage{subfigure}
\usepackage{graphicx}
\usepackage{subcaption}

\usepackage{algorithm}
\usepackage{algpseudocode}

\usepackage{caption}

\usepackage[T1]{fontenc}

\usepackage[utf8]{inputenc}

\usepackage{microtype}
\usepackage{multicol}
\usepackage{booktabs}
\usepackage{multirow}
\usepackage{xcolor}         
\usepackage{enumitem}
\usepackage{listings}
\usepackage{colortbl}
\usepackage{amsmath}
\usepackage{amssymb}
\usepackage{inconsolata}

\usepackage{graphicx}

\newcommand\logo{\raisebox{-12pt}{\includegraphics[width=3.0em]{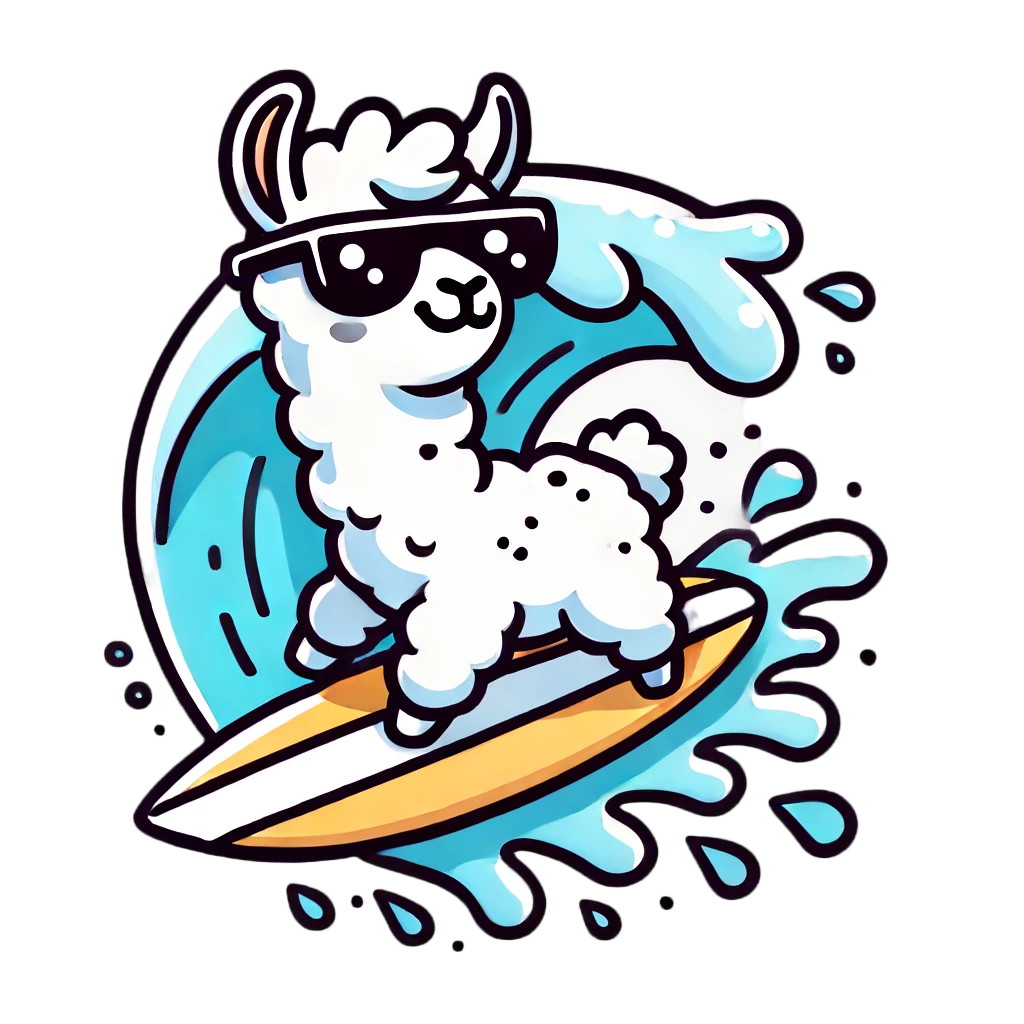}}}

\title{\logo{} SURf: Teaching Large Vision-Language Models to Selectively Utilize Retrieved Information}

\author{
Jiashuo Sun\textsuperscript{\rm 1}\thanks{Work done during internship at Shanghai AI Laboratory.},
Jihai Zhang\textsuperscript{\rm 2},
Yucheng Zhou\textsuperscript{\rm 3}, 
\textbf{
Zhaochen Su\textsuperscript{\rm 4},
Xiaoye Qu\textsuperscript{\rm 5}\textsuperscript{†},
Yu Cheng\textsuperscript{\rm 2}\thanks{Both are corresponding authors.}}
\\
\textsuperscript{\rm 1} 
Xiamen University,
\textsuperscript{\rm 2}
The Chinese University of Hong Kong\\
\textsuperscript{\rm 3}
SKL-IOTSC, CIS, University of Macau, \textsuperscript{\rm 4}
Soochow University \\
\textsuperscript{\rm 5}
Shanghai AI Laboratory
}

\begin{document}
\maketitle

\begin{abstract}

Large Vision-Language Models (LVLMs) have become pivotal at the intersection of computer vision and natural language processing. However, the full potential of LVLMs' Retrieval-Augmented Generation (RAG) capabilities remains underutilized. Existing works either focus solely on the text modality or are limited to specific tasks. Moreover, most LVLMs struggle to selectively utilize retrieved information and are sensitive to irrelevant or misleading references. To address these challenges, we propose a self-refinement framework designed to teach LVLMs to \textbf{S}electively \textbf{U}tilize \textbf{R}etrieved In\textbf{f}ormation (SURf). Specifically, when given questions that are incorrectly answered by the LVLM backbone, we obtain references that help correct the answers (positive references) and those that do not (negative references). We then fine-tune the LVLM backbone using a combination of these positive and negative references. Our experiments across three tasks and seven datasets demonstrate that our framework significantly enhances LVLMs’ ability to effectively utilize retrieved multimodal references and improves their robustness against irrelevant or misleading information. The source code is available at {\url{https://github.com/GasolSun36/SURf}}.
\end{abstract}

\section{Introduction}

Large Vision-Language Models (LVLMs) have become crucial at the intersection of computer vision and natural language processing (NLP), empowering various applications by generating contextually relevant textual descriptions from visual inputs \cite{llava-1.5, gpt4v, instructblip, qwen-vl, mplug-owl, minigpt4, fan2024nphardeval4v, sun2024visual}. These models capture and translate complex visual patterns into coherent linguistic representations. The development of LVLMs is driven by continuous improvements in model architecture, training methodologies, and data diversity \cite{wang2024mitigating, wang2024vigc, yu2023rlhf,qu2024mitigating}, resulting in better performance and broader applicability.

\begin{figure}[t]
\centering
\includegraphics[width=1.0\linewidth]{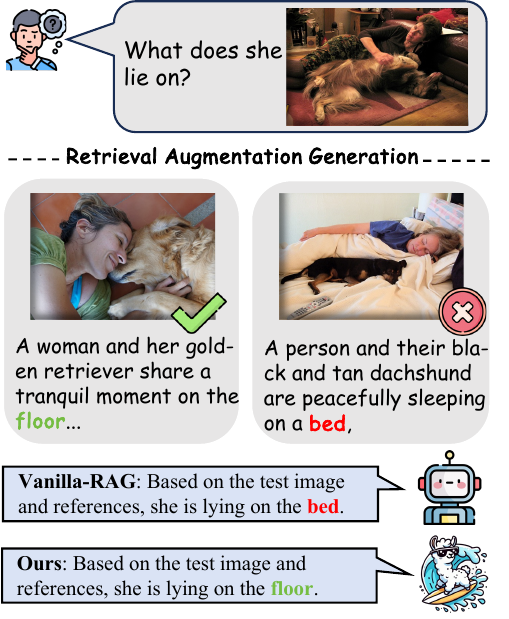}
\caption{Illustration of multimodal RAG. RAG can introduce misleading content, causing LVLMs to generate incorrect responses. SURf can selectively utilize information from images and descriptions, e.g., the first image-caption pair.).}
\label{fig:introduction}
\end{figure}

\begin{figure*}[t]
\centering
\includegraphics[width=\linewidth]{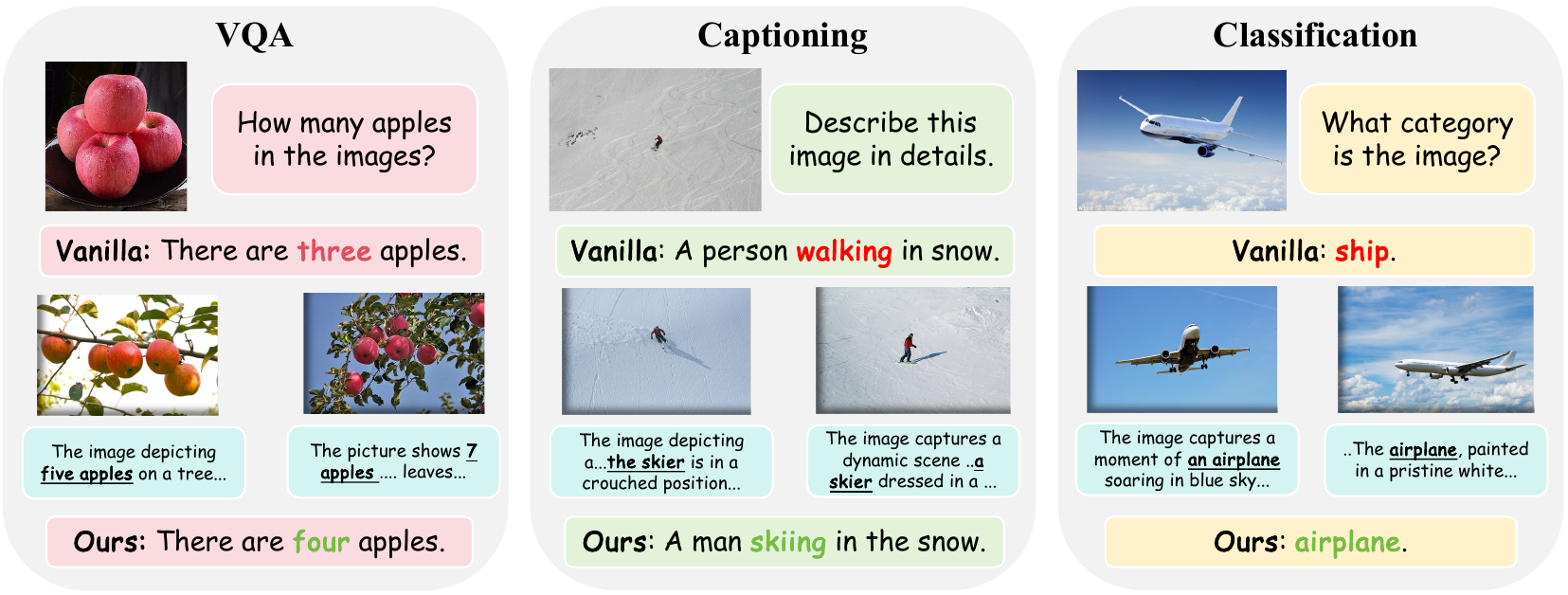}
\caption{The illustration of Multimodal RAG for VQA, Captioning and Classification Tasks. Providing images similar to the test images along with their descriptions as references can help LVLMs answer questions more accurately.}
\label{fig:three_tasks}
\end{figure*}

Although LVLMs excel in visual language representation, they struggle with image generalization and understanding \cite{qu2024look}. Similarly, LLMs face these challenges in the NLP domain but can mitigate them by incorporating additional knowledge or references through Retrieval-Augmented Generation (RAG), ensuring high trustworthiness \cite{dpr, self-rag, recomp}. However, in LVLMs, the full potential of RAG remains under-explored. Firstly, many previous multimodal RAG-related works have only focused on the text modality \cite{samll_cap, rag_captioning, lmcap}, without fully utilizing the LVLMs' understanding of visual content. Secondly, the few works that integrate multimodal references are often limited to specific tasks like image captioning, ignoring the broader potential of applying RAG technology \cite{zero_shot_captioning, rag_modeling}. 
Finally, a significant issue overlooked by existing research is the potential irrelevance or even disruptive nature of retrieved content in practical applications.
Under this circumstance, vanilla LVLMs fail to dynamically select retrieval content, but treat them indiscriminately, leading to a performance decline \cite{ra-dit,qu2024alleviating}. 

In this paper, we propose a self-refinement framework that enables LVLMs to selectively utilize the retrieved information from both image and text sources while effectively enhancing the model's robustness against irrelevant or misleading content. Specifically, we identify the visual questions that are wrongly answered by LVLM and use image-caption pairs to prompt the LVLMs to generate responses. Secondly, we assess the contribution of the introduced image-caption pairs by invoking external evaluation tools, thereby constructing a training dataset with positive and negative samples. Subsequently, we build a RAG instruction dataset to further train the LVLMs, allowing them to better benefit from RAG tasks and improve their robustness against irrelevant retrieval content. It is worth noting that we only reconstruct data from the SFT phase of the LVLMs without using any additional new datasets. We extensively evaluate our method across seven datasets and benchmarks in three different tasks: VQA, image captioning, and image classification. The experimental results demonstrate that our approach can further enhance the RAG capabilities of existing LVLMs and significantly improve their robustness in generating responses when faced with irrelevant images or content.

Our contributions are summarized as follows: (1) We empirically demonstrate that integrating Multimodal RAG with LVLMs can improve model performance, while also revealing that current LVLMs are highly sensitive to irrelevant and misleading retrieval information, which presents a significant challenge. (2) We design a lightweight and cost-effective self-refinement framework specifically aimed at teaching LVLMs to selectively utilize relevant information. (3) Through extensive experiments and evaluations, we show that our approach enhances the models' ability to effectively utilize retrieval information, making them more robust against irrelevant and misleading references.

\begin{figure*}[htbp]
\centering
\includegraphics[width=\linewidth]{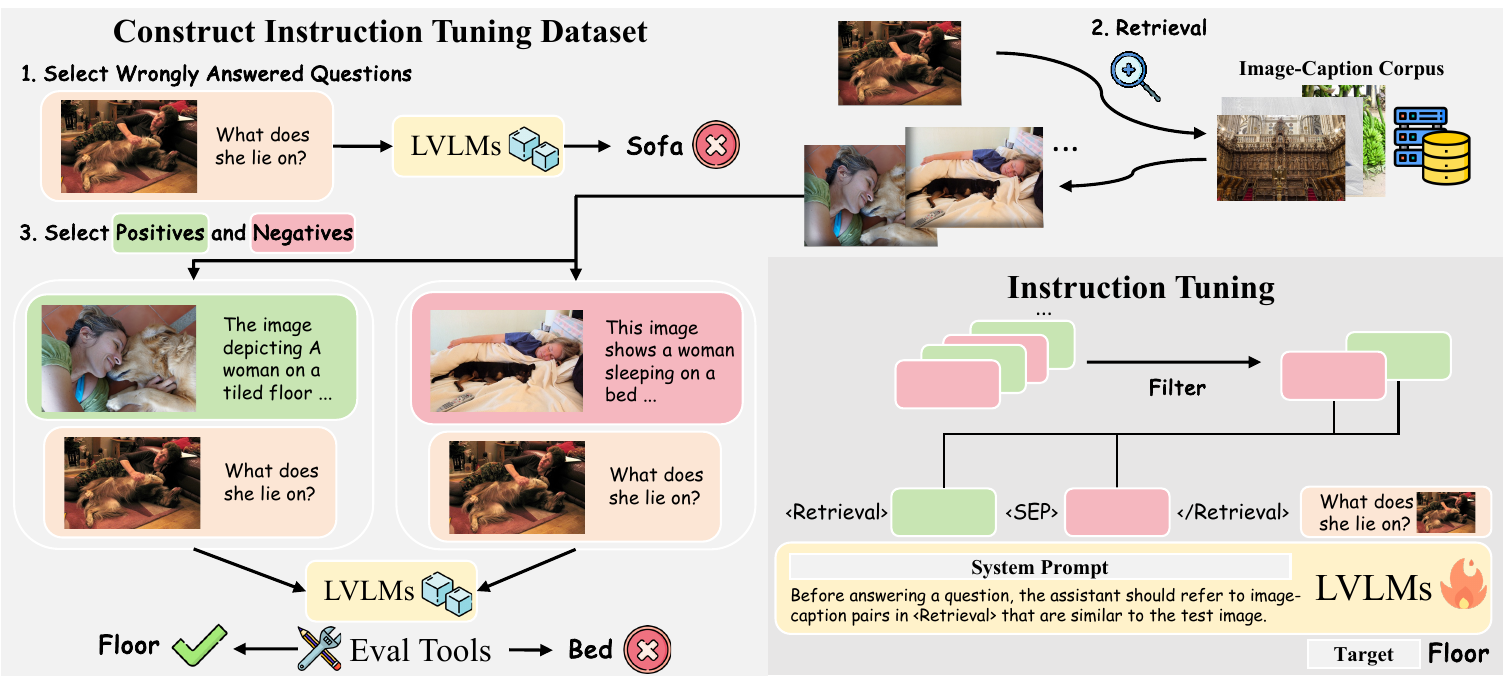}
\caption{Illustration of our training framework. First, we collect questions that LVLMs initially answered incorrectly. Next, we retrieve the Top-N image-caption pairs from the corpus, allowing the LVLM to reattempt the questions. We then evaluate the answers to see if they have improved (positive) or worsened (negative). After that, we filter for the highest-quality training data and use it for instruction tuning to train the LVLMs.}
\label{fig:pipeline}
\end{figure*}

\section{Robust Multimodal RAG}

\subsection{Preliminaries}


The RAG consists of two main components: a retriever and a generator. The retriever fetches relevant information from a large document collection, and the generator uses the retrieved document to produce the final output. We can represent the functioning of the RAG in LVLMs with the following formulas: 

Given an input \( x \) (e.g., a question $q$ or instruction with a feature vector of an image $\text{i}_{\text{test}}$), the retriever fetches \( k \) relevant images \( \{i_1, i_2, \ldots, i_k\} \) from an image set of image-caption collection $\mathbb{D}$. 
The probability distribution of the retriever can be represented as $\bar{P}(i \mid x)$. 
The generator uses the retrieved images \( \{i_1, i_2, \ldots, i_k\} \), the corresponding captions \( \{c_1, c_2, \ldots, c_m\} \) and the input \( x \) to generate the output \( y \) (e.g., an answer, image caption, or classification label). The conditional probability distribution of the generator can be represented as: 
\begin{align}
P(y \mid x, \{[i_1, c_1], [i_2, c_2] \ldots, [i_k, c_k]\}
\end{align}

The final output of the LVLM with RAG is based on the joint probability of the input \( x \) and the set of retrieved image-caption pairs \( \{[i_1, c_1], [i_2, c_2], \ldots, [i_k, c_k]\} \):
\begin{align}
    P(y \mid x) = \sum_{j=1}^k P(y \mid x, r_j) \bar{P}(i_j \mid x) 
\label{eq:1}
\end{align}
where $r_i = [i_j, c_j]$ represents the retrieved image-caption pair, with $c_j$ being the caption corresponding to the retrieved image $i_j$.


\begin{figure}[t]
\centering
\includegraphics[width=1.0\linewidth]{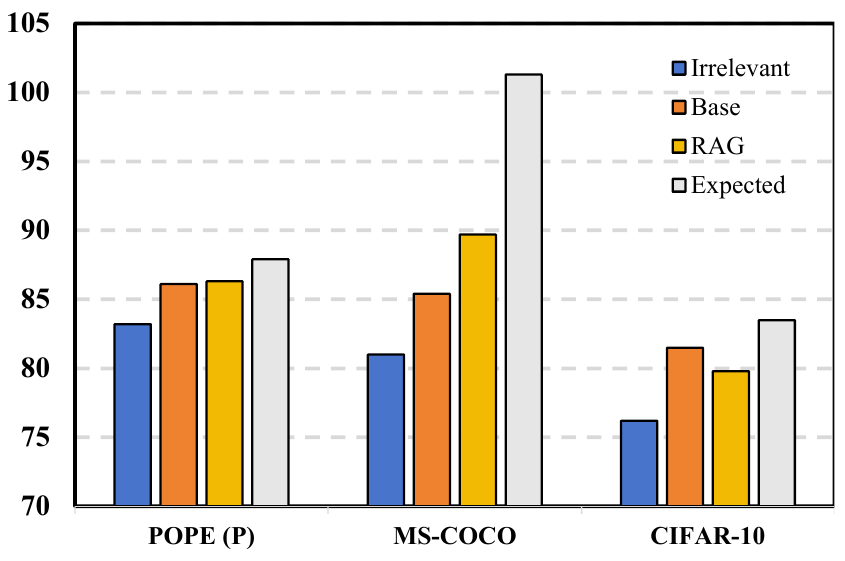}
\caption{Performance of the base model (LLaVA-1.5-7B) without using RAG (Base), RAG with irrelevant content (Irrelevant), and RAG on POPE-popular, MS-COCO, and CIFAR-10.}
\label{fig:expected_results}
\end{figure}

\subsection{Multimodal RAG Benefit LVLMs}
RAG has been proven to improve model performance on downstream tasks while maintaining a high level of trustworthiness in the field of NLP \cite{dpr, self-rag, recomp, jin2024impact}. However, in LVLMs, the full potential of RAG remains under-explored.

As shown in Figure \ref{fig:three_tasks}, when addressing tasks such as VQA, captioning, and classification, we can enhance the model performance by retrieving relevant images and their corresponding descriptions to provide a pattern mapping for the input $x$. The collection of pattern state is denoted as $\mathbb{M} = \{M_0, M_1, \cdots, M_n\}$, and 
$M_i \sim ([I,T] \in \mathbb{M})$
, where $I$ and $T$ denote the image and description, respectively. Next, our goal is to learn this mapping:
\begin{align}
f: x \rightarrow f(x|M_{i_1}, M_{i_2}, \ldots, M_{i_k})
\end{align}

To better understand the impact of RAG on model performance, we conducted experiments comparing direct inference with RAG-enhanced inference across three datasets. Figure \ref{fig:expected_results} illustrates the performance differences. It can be observed that retrieving and incorporating additional multimodal information (both image and text) significantly improves the model's performance in tasks across VQA, Image Captioning, and Classification.

\subsection{Irrelevant Harms Model Performance}
Typically, the retrieval process \( \bar{P}(i_j \mid x) \) or \( \bar{P}(c_j \mid x) \) is typically implemented by computing image-to-image or image-to-text similarity in CLIP embedding space \cite{samll_cap, rag_captioning}. However, this retrieval process is not always reliable, leading to the inclusion of irrelevant or misleading references. For example in Figure \ref{fig:introduction}, the similarity scores returned the \text{Top-2} images most similar to the test image. Nevertheless, these two images contribute differently to the original question. The latter image misleads the model and causes incorrect responses.

Figure \ref{fig:expected_results} demonstrates the impact of irrelevant information on RAG. It can be seen that the performance of RAG is even worse than without introducing any additional information, which indicates the negative impact of irrelevant or disturbing information on current LVLMs. We believe that the RAG of current LVLMs still has significant potential. If we can teach the model to selectively utilize the retrieved information and ignore the irrelevant or misleading ones, the performance of RAG in LVLMs will be further improved, potentially approaching the results shown by the gray bars.





\subsection{Robust RAG Training Framework}
Since RAG has great potential to help improve the accuracy of model generation, and regardless of how the retriever is optimized, achieving perfect retrieval recall is unattainable \cite{clip, open-clip}. Therefore, we choose to optimize $P(y \mid x, r_i),$ through teaching the model to learn to selectively utilize the retrieved information. We propose a self-refinement framework that enables LVLMs to selectively refer to relevant information from both image and text sources while effectively enhancing the model's robustness against irrelevant or misleading content.

\begin{table*}[t]
\resizebox{\textwidth}{!}{
\begin{tabular}{lcccccccccc}
\toprule
            & \multicolumn{5}{c}{\textbf{VQA}}                                                        & \multicolumn{2}{c}{\textbf{Captioning}}  & \multicolumn{2}{c}{\textbf{Classification}}   & \multirow{2}{*}{\textbf{Avg.}} \\
            & POPE (R)   & POPE (P)  & POPE (A) & MMstar        & $\text{Vizwiz}^{V}$ & MS-COCO &    $\text{Vizwiz}^{C}$ & CIFAR-10 & EmoSet &  \\ \midrule
\multicolumn{11}{c}{\cellcolor{gray!20}\textit{\textbf{7B Parameter Model}}}                                                                                                                                               \\
Zero-shot   & 87.3          & 86.1          & 84.2           & 30.3          & 50.0          & 198.6          & 134.5          & 81.5     & 52.8   & 89.48              \\
Vanilla-RAG & 87.9          & 86.3          & 83.3           & 32.1          & 48.3          & 178.1          & 169.6          & 79.7     & 50.4   & 90.63              \\
Rerank-RAG  & 88.3          & 86.3          & 83.4           & 31.4          & 49.3          & 210.0          & 164.2          & 80.9     & 50.5   & 93.81              \\
Filter-RAG  & 88.5          & 86.6          & \textbf{83.9}  & 31.8          & 51.5          & 231.1          & 172.0          & 82.2     & 51.8   & 97.71              \\
\textbf{SURf} & \textbf{89.8} & \textbf{87.9} & 83.6           & \textbf{33.5} & \textbf{54.3} & \textbf{238.4} & \textbf{177.4} & \textbf{83.5} & \textbf{53.1} & \textbf{100.17} \\
\multicolumn{11}{c}{\cellcolor{gray!20}\textit{\textbf{13B Parameter Model}}}                                                                                                                                              \\
Zero-shot   & 87.1          & 86.2          & 84.5           & 32.8          & 53.6          & 210.0          & 150.2          & 82.6     & 56.4   & 93.71              \\
Vanilla-RAG & 88.3          & 86.4          & 83.4           & 33.1          & 50.2          & 218.5          & 160.9          & 80.7     & 55.6   & 95.23              \\
Rerank-RAG  & 88.4          & 86.4          & 83.6           & 33.5          & 50.9          & 223.1          & 162.1          & 82.0     & 56.0   & 96.22              \\
Filter-RAG  & 88.6          & 86.5          & 83.8           & 33.7          & 51.7          & 226.7          & 164.1          & 83.2     & 56.5   & 97.20              \\
\textbf{SURf} & \textbf{89.5} & \textbf{87.7} & \textbf{84.6}  & \textbf{34.5} & \textbf{54.6} & \textbf{250.9} & \textbf{177.5} & \textbf{85.1} & \textbf{58.1} & \textbf{102.50} \\ \bottomrule
\end{tabular}
}
\caption{Performance comparison of our model on 7B and 13B parameters using four methods across seven tasks. In POPE, (R), (P), and (A) stand for Random, Popular, and Adversarial subsets, respectively (applies to all tables below.). $\text{Vizwiz}^{V}$ and $\text{Vizwiz}^{C}$ represents VQA and captioning based on Vizwiz. The best performance in the table is highlighted in \textbf{bold}.}
\label{tab:main_results}
\end{table*}

\begin{table*}[]
\resizebox{\textwidth}{!}{
\begin{tabular}{lccccccccc}
\toprule
              & \multirow{2}{*}{Para.} & \multirow{2}{*}{Shots} & \multicolumn{2}{c}{\textbf{POPE (R)}} & \multicolumn{2}{c}{\textbf{POPE (P)}} & \multicolumn{2}{c}{\textbf{POPE (A)}} & \textbf{MS-COCO} \\
              &                        &                        & Acc.         & F1           & Acc.         & F1           & Acc.         & F1           & CIDEr $\uparrow$   \\ \midrule
Flamingo \cite{flamingo} & 9B                     & 4-shots                & -         & -         & -         & -         & -         & -         & 93.1    \\
OpenFlamingo \cite{openflamingo} & 9B                     & 4-shots                & 48.5         & 48.1         & 49.5         & 49.0         & 48.9         & 48.5         & 89.0    \\
Otter \cite{otter}         & 9B                     & 4-shots                & 82.5         & 81.8         & 74.7         & 73.9         & 69.9         & 69.4         & 92.2    \\
MMICL \cite{MMICL}         & 12.1B                  & 4-shots                & 87.3         & 86.6         & 82.7         & 82.1         & 81.0         & 80.7         & 95.7    \\
\textbf{SURf}          & 7B                     & 2-shots                & \textbf{89.8}         & \textbf{89.3}         & \textbf{87.9}         & \textbf{87.6}         & \textbf{83.6}         & \textbf{83.9}         & \textbf{101.3}   \\ \bottomrule
\end{tabular}
}
\caption{Performance of our 7B model compared to four ICL models on the three POPE subsets (VQA) and MS-COCO (captioning). The results of the ICL models are directly from the original paper.}
\label{tab:icl_models}
\end{table*}

\subsubsection{Construction of Positive and Negative Examples}

Introducing both relevant and irrelevant content during training can enhance the model's ability to distinguish and select relevant information \cite{ra-dit}. Therefore, at this stage, we construct positive and negative examples (denoted as $\mathbb{C}_{\text{pos}}$ and $\mathbb{C}_{\text{neg}}$) for subsequent robust training.

We hypothesize that if the model initially answers a question incorrectly but can answer correctly after including an example (both image and description), that example contains useful information (positive). Otherwise, the example is considered misleading or irrelevant (negative). Specifically, we first collect the data used by the LVLM during the SFT stage and use a fixed-parameter LVLM to answer questions based on images, recording incorrect examples. Then, we perform retrieval from the image-caption corpus to obtain the \text{Top-N} images and their corresponding descriptions. These are then appended to the test image and question, allowing the LVLM to answer the question again. We use specific evaluation tools to determine whether the answer has improved, remained unchanged, or worsened. Image-caption pairs that successfully improve the answer are considered positive examples of the current question, while those that do not cause any change or worsen the answer are considered negative examples of the current question.

Notably, the data we construct is sourced from the examples in the existing LVLM training data used during the instruction fine-tuning stage, requiring no new external data.

\subsubsection{Data Filtering}
\label{sec:data_filtering}

Due to the token length limitation in LVLMs, we need to further filter the positive and negative examples obtained in the previous step. We exclude examples from the \text{Top-N} image-caption pairs that contain only positive or negative examples. For positive examples, we select the image with the highest similarity to the test image to ensure the inclusion of highly relevant information and to avoid model training collapse:
\begin{align}
p_{\text{pos}} \sim \max_{i_j \in \mathbb{C}_{\text{pos}}} p_V^\theta(x, i_j)
\end{align}

For negative examples, we choose the image with the highest similarity to the test image as hard negatives. These hard negatives are more similar to the positive examples, thus requiring the model to develop higher discriminative capabilities to accurately identify them:
\begin{align}
p_{\text{neg}} \sim \max_{i_j \in \mathbb{C}_{\text{neg}}} p_V^\theta(x, i_j)
\end{align}

\subsubsection{RAG Instruction-Tuning}
Using the high-quality positive and negative example pairs generated through the above process, we fine-tune the existing model with RAG instructions. The retrieved images and their corresponding descriptions are concatenated sequentially before the test image, enclosed by special characters \texttt{<Retrieval>} and \texttt{</Retrieval>}. This ensures that the model can effectively distinguish between retrieved-context and the actual test input, enhancing its ability to leverage relevant information while minimizing the impact of irrelevant or misleading data.

The algorithm of our method is shown in the Appendix Algorithm~\ref{algorithm}.

\begin{table*}[t]
\resizebox{\textwidth}{!}{
\begin{tabular}{lcccccccccccc}
\toprule
            & \multicolumn{2}{c}{\textbf{\textbf{POPE (R)}}} & \multicolumn{2}{c}{\textbf{\textbf{POPE (P)}}} & \multicolumn{2}{c}{\textbf{\textbf{POPE (A)}}} & \multicolumn{5}{c}{\textbf{\textbf{MS-COCO}}}             & \textbf{CIFAR-10} \\
            & Acc.         & F1           & Acc.         & F1           & Acc.         & F1           & B@4  & METEOR & ROUGE-L & CIDEr & SPICE & Acc.     \\ \midrule
\textbf{Zero-shot}   & 87.3         & 86.0         & 86.1         & 84.9         & 84.2         & 83.4         & 22.3 & 28.0   & 50.9    & 75.3  & 22.1  & 81.5     \\ \midrule
\textbf{Vanilla-RAG} & \textbf{87.9} & \textbf{86.5} & \textbf{86.3} & \textbf{85.0} & \textbf{83.3} & \textbf{82.5} & \textbf{24.5} & \textbf{28.3} & \textbf{51.4} & \textbf{79.8} & \textbf{22.4} & \textbf{79.7} \\
w/ 1k & 87.5 & 86.3 & 85.4 & 84.2 & 82.2 & 81.3 & 22.5 & 27.8 & 50.5 & 75.0 & 21.8 & 79.5 \\
w/ 5k & 87.4 & 86.2 & 85.3 & 84.1 & 82.2 & 81.3 & 22.2 & 27.7 & 50.4 & 75.0 & 21.6 & 77.1 \\
w/ 10k & 87.2 & 86.0 & 85.0 & 83.8 & 82.1 & 81.2 & 22.0 & 27.4 & 50.1 & 75.4 & 21.2 & 76.7 \\
w/ 100k & 87.0 & 85.9 & 84.9 & 83.7 & 82.0 & 81.1 & 22.1 & 27.3 & 50.3 & 74.8 & 21.5 & 76.4 \\
w/ 1,000k & 86.7 & 85.0 & 84.5 & 83.1 & 81.8 & 80.8 & 22.0 & 27.5 & 49.9 & 73.6 & 21.2 & 75.6 \\ \midrule
\textbf{Ours} & \textbf{89.8} & \textbf{89.3} & \textbf{87.9} & \textbf{87.6} & \textbf{83.6} & \textbf{83.9} & \textbf{27.9} & \textbf{29.9} & \textbf{55.1} & \textbf{101.3} & \textbf{24.2} & \textbf{83.5} \\
w/ 1k & 88.9 & 88.3 & 87.8 & 87.3 & 83.1 & 83.0 & 26.8 & 29.4 & 54.2 & 97.3 & 23.7 & 83.1 \\
w/ 5k & 89.3 & 88.7 & 87.8 & 87.3 & 83.1 & 83.4 & 26.5 & 29.1 & 53.6 & 97.4 & 23.5 & 82.4 \\
w/ 10k & 89.4 & 88.8 & 87.6 & 87.3 & 83.2 & 83.5 & 26.9 & 29.4 & 54.2 & 97.7 & 23.8 & 83.4 \\
w/ 100k & 89.1 & 88.5 & 87.7 & 87.2 & 83.3 & 83.4 & 26.6 & 29.2 & 53.9 & 96.5 & 23.6 & 80.5 \\
w/ 1,000k & 89.2 & 88.7 & 87.9 & 87.4 & 83.6 & 83.6 & 27.1 & 29.4 & 54.3 & 98.4 & 23.7 & 80.9 \\  \bottomrule
\end{tabular}
}
\caption{Performance comparison of our model and vanilla-RAG on three tasks when introducing irrelevant image-caption pairs. "1k to 1,000k" indicates the range of similarity between the introduced images and the test images, with larger values indicating less relevance.}
\label{tab:irrelevant1}
\end{table*}

\begin{table*}[t]
\resizebox{\textwidth}{!}{
\begin{tabular}{lcccccccccccc}
\toprule
                   & \multicolumn{2}{c}{\textbf{\textbf{POPE (R)}}} & \multicolumn{2}{c}{\textbf{\textbf{POPE (P)}}} & \multicolumn{2}{c}{\textbf{\textbf{POPE (A)}}} & \multicolumn{5}{c}{\textbf{MS-COCO}}             & \textbf{CIFAR-10} \\
                   & Acc.         & F1           & Acc.         & F1           & Acc.         & F1           & B@4  & METEOR & ROUGE-L & CIDEr & SPICE & Acc.     \\ \midrule
Vanilla-RAG        & 87.9         & 86.5         & 86.3         & 85.0         & 83.3         & 82.5         & 24.9 & 28.5   & 52.8    & 89.7  & 22.6  & 79.5     \\
w/ Switch & 87.2         & 86.0         & 85.7         & 84.6         & 82.4         & 82.0         & 22.2 & 28.0   & 50.8    & 75.1  & 22.0  & 78.4     \\ \midrule
Ours               & 89.8         & 89.3         & 87.9         & 87.6         & 83.6         & 83.9         & 27.9 & 29.9   & 55.1    & 101.3 & 24.2  & 83.5     \\
w/ Switch        & 89.6         & 89.1         & 87.9         & 87.6         & 83.6         & 83.8         & 26.8 & 29.3   & 54.1    & 97.1  & 23.7  & 83.4     \\ \bottomrule
\end{tabular}
}
\caption{Performance comparison of our model and vanilla-RAG on three tasks in the random switching of retrieved content setting.}
\label{tab:irrelevant2}
\end{table*}

\section{Experiment}


\subsection{Datasets}
We evaluated our model using seven datasets across three distinct tasks: VQA: POPE \cite{pope}, MMStar \cite{mmstar}, Vizwiz-VQA \cite{vizwiz_vqa}, Image Captioning: MS-COCO \cite{ms-coco}, Vizwiz-Caption \cite{vizwiz-caption}, Image Classification: CIFAR-10 \cite{cifar-10}, EmoSet \cite{emoset}. For more detailed information and metrics can be found in the Appendix \ref{sec: data_analysis}.

\subsection{Baselines}
We compared four methods among LlaVA-1.5-7B and LLaVA-1.5-13B \cite{llava-1.5}:

\noindent\textbf{Zero-shot} Directly prompting LVLMs to generate responses.

\noindent\textbf{Vanilla-RAG} Concatenating the Top-N image-caption pairs from the database, which have the highest CLIP score similarity to the test image before the questions and images for the LVLMs to respond. 

\noindent\textbf{Rerank-RAG} Building on Vanilla-RAG, we prompt the LLM to generate a caption for the test image. We then calculate the BERT-Score between this caption and the descriptions of the retrieved images, ranking the image-caption pairs with higher relevance scores at the top.

\noindent\textbf{Filter-RAG} Enhancing Rerank-RAG by removing any image-caption pairs with a similarity score less than $S$.

Additionally, we compared four In-Context Learning (ICL) models: Flamingo \cite{flamingo}, OpenFlamingo \cite{openflamingo}, Otter \cite{otter}, and MMICL \cite{MMICL}. For all approaches, we used greedy decoding as the decoding strategy.

\subsection{Implementation details}
We collected 60,000 initial incorrect responses from LVLMs and generated 10,000 samples with positive and negative sample pairs. After filtering, we refined this to 2,000 samples for the final training data. We use LLaVA-1.5 as the LVLM backbone of our model SURf-7B and SURf-13B and use CLIP (ViT-L with a resolution of 336*336) \cite{clip} as the vision encoder. Our 7B and 13B models are further trained from the instruction-finetuned LLaVA-1.5-7B and LLaVA-1.5-13B models following previous works \cite{moe-llava, video-llava, llava-med, llava-plus} since LLaVA is the most popular used LVLMs. We use 8 A100-80G to training 1 hour for 2 epochs. For the VQA and image captioning task, we use exact match and Bert-Score \cite{bert-score} as the evaluation tool respectively, mentioned in Section \ref{sec:data_filtering}. We use ShareGPT4v-PT \cite{sharegpt4v} as our database for RAG, which includes approximately 1,246k image-caption pairs with an average caption length of 826. For the retrieval system, we use FAISS \cite{faiss} with flat indexes to pre-index the computed embeddings of all images in the database.

\begin{figure*}[htbp]
    \centering
    \subfigure[Performance of our model using different retrieval sources.]{
        \includegraphics[width=0.45\textwidth]{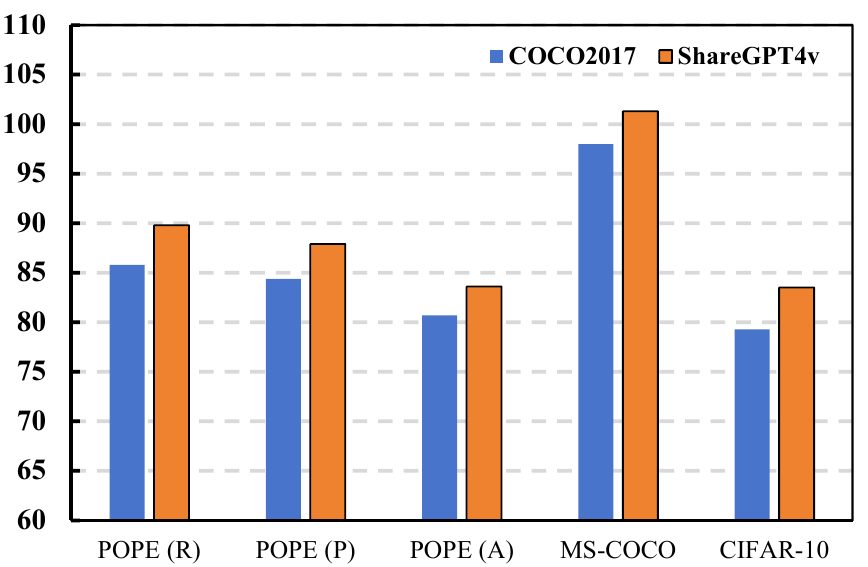}
        \label{fig:database}
    }
    \hfill
    \subfigure[Performance of our model on downstream tasks with and without data filtering.]{
        \includegraphics[width=0.45\textwidth]{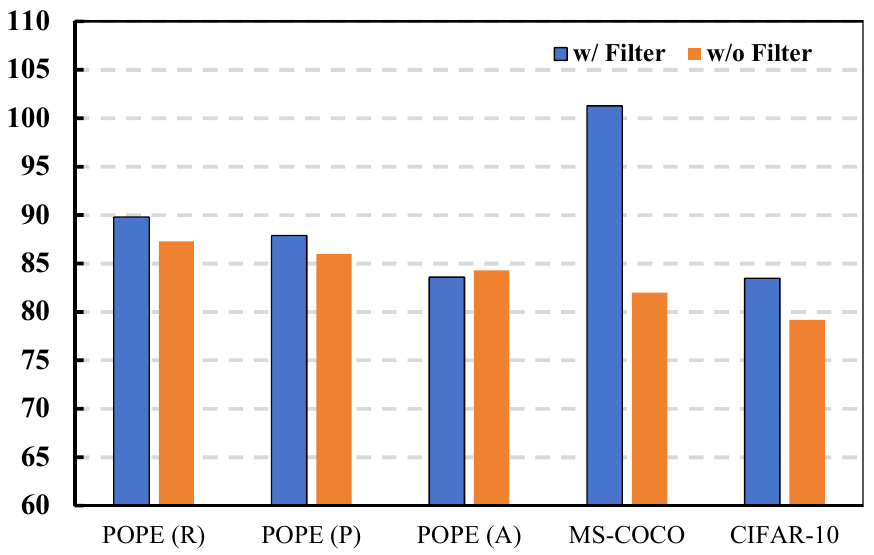}
        \label{fig:filter}
    }
    \caption{Ablation Study of Database Size and Data Filter.}
    \label{fig:example}
\end{figure*}

\begin{figure}[htbp]
\centering
\includegraphics[width=1.0\linewidth]{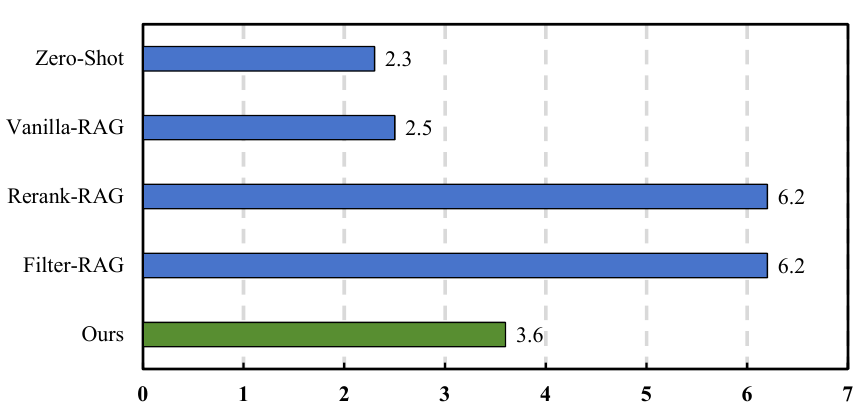}
\caption{Efficiency analysis of our model compared to four methods. We report the running times per sample.}
\label{fig:efficiency_analysis}
\end{figure}

\subsection{Experimental Results}

\paragraph{Compare to Baselines}
Table \ref{tab:main_results} presents the comparison of our model, trained with our method, against four other methods. On the VQA task, our model significantly outperforms previous methods, with a VQA accuracy improvement of approximately 3.7\% for the 7B model compared to zero-shot and 2.3\% compared to Filter-RAG, achieving state-of-the-art results. Furthermore, on the captioning task, the improvement of our model is even more pronounced (detailed results can be found in the Appendix).

In contrast, for the classification task, vanilla-RAG may perform worse than direct inference. However, our training method enables the model to selectively refer to the retrieved content, resulting in a final performance that significantly surpasses zero-shot. For the 13B model, the improvement in captioning is even more significant, with an approximate 34.1\% increase compared to zero-shot. Additionally, the table illustrates that simple methods, such as reranking and filtering, cannot effectively address the problem of irrelevant content introduced by retrieval.

\paragraph{Compare to ICL models}
The experiments in Table \ref{tab:icl_models} compare our model with various ICL models, as ICL models are very similar to ours at the input level. Despite having fewer parameters and exemplars (For the ICL models, more shots correspond to better performance, we used their 4-shot results for comparison since they only reported results for 4-shot or 32-shot scenarios.) in the prompts compared to the other models, our model achieves the best performance on both the POPE and MS-COCO datasets. Specifically, it improves the average accuracy by 3.4\% and the F1 score by 3.8\% on POPE compared to the second-best model. This demonstrates that our model can effectively utilize the retrieved content to enhance the performance of downstream tasks.

\paragraph{Robustness}
Tables \ref{tab:irrelevant1} and \ref{tab:irrelevant2} present the results of our robustness tests. In Table \ref{tab:irrelevant1}, we maintain the image-caption pair with the highest CLIP similarity score among the retrieved content to ensure effective information. We also introduce image-caption pairs from the Top-K (from 1k to 1,000k) positions as forced irrelevant information. The results show that the performance of vanilla-RAG significantly declines on the three datasets as more irrelevant image-caption pairs are introduced. In contrast, our model's performance remains very stable. Notably, the model's performance when introducing 100k and 1000k irrelevant image-caption pairs is better than when introducing 1k pairs. This improvement is because, after training with hard negative samples, our model can easily distinguish content unrelated to the test image and question, thereby focusing more on other relevant information in the retrieval.

Table \ref{tab:irrelevant2} shows that our model remains robust even after randomly shuffling the examples, whereas vanilla-RAG exhibits a significant decline in performance. This demonstrates that training the model with our proposed framework enables it to selectively extract relevant information from the retrieved content, making it less sensitive to the order of the examples.

\subsection{Ablation Study}

\paragraph{Size of the Database}
We conducted experiments using different databases as retrieval sources, with results shown in Figure \ref{fig:database}. It can be seen that using COCO-2017 (approximately 118k image-caption pairs) and ShareGPT-4V (approximately 1,246k image-caption pairs) results in notable differences in model performance for VQA and classification tasks, while the metrics for the captioning task show minimal differences. The reason is that VQA and classification tasks are more challenging for the model compared to captioning, requiring a larger retrieval source to provide more diverse reference image-caption pairs.

\paragraph{Data Filter}
Figure \ref{fig:filter} presents the results with and without using the data filtering step. It can be seen that the performance of the model trained without data filtering is significantly worse compared to when data filtering is used. This highlights the importance of filtering positive and negative samples and training with hard negative sampling in our training framework.

\paragraph{Efficiency Analysis}
We compared the efficiency of our method with four other methods, as shown in Figure \ref{fig:efficiency_analysis}. We calculate the average running time for 1,000 samples in image captioning tasks with the max token length set to 256. It can be observed that our method increases the time by approximately 1.3 seconds per sample compared to the zero-shot approach. This increase is primarily due to the time required to convert the image to an embedding, retrieval time, and the additional overhead introduced by the increased length of the prompt. However, this slight increase in time is acceptable considering the performance improvement.

In contrast, Rerank-RAG and Filter-RAG are slower because they require additional prompts for the LVLMs to generate captions for the current image, which are then used for text similarity comparisons.

\section{Related Work}
\subsection{Large Vision-Language Models}
Large vision-language models (LVLMs) have greatly benefited from advancements in large language models (LLMs) \cite{llama2, vicuna, su2024living, su2024timo}, which integrate a vision encoder with a language model backbone. Leveraging the success of LLMs through pre-training and instruction tuning \cite{llava-1.5, mplug-owl, minigpt4, qwen-vl, instructblip}, LVLMs like LLaVA \cite{llava-1.5} employ GPT-4V\footnote{https://openai.com/research/gpt-4v-system-card} to generate diverse instruction datasets, thereby enhancing their capacity to understand images and follow human instructions. Despite these successes, current MLLMs still face significant challenges with hallucinations \cite{su2022improving, li2023evaluating, liu2023mitigating, wang2024vigc, zhao2024mitigating, leng2023mitigating,zhang2024avoiding,ZhouLWS24}. These issues often result from misalignment between the vision and language components, leading to neglecting image details and generating incorrect content. Our work aims to improve LVLMs' ability to selectively reference retrieved information and enhance robustness against misleading content.


\subsection{Retrieval-Augmented Generation}
Retrieval-augmented generation (RAG) has become a powerful approach in natural language processing, combining the strengths of retrieval-based methods and generative models \cite{merth2024superposition, self-rag, recomp, ra-dit, su2024conflictbank}. In the NLP domain, RAG aims to select the most relevant documents from a large corpus using techniques such as BM25 \cite{bm25} and neural retrievers like DPR \cite{dpr}. However, the challenge in the multimodal domain is considerably higher, as the retrieval dimension encompasses images along with text. Previous works \cite{zero_shot_captioning, samll_cap, rag_modeling, rag_captioning, lmcap} have shown that retrieving similar images based on a test image and using their corresponding captions can enhance model performance on captioning tasks. Nevertheless, these methods often fail to address how to manage irrelevant image-caption pairs, which can decrease model accuracy. Our work focuses on improving LVLMs' ability to selectively reference pertinent retrieved information and increase robustness against misleading content, thereby enhancing performance across various downstream tasks.



\section{Conclusion}

This paper introduces a robust self-refinement multimodal RAG training framework designed for LVLMs. Our approach incorporates retrieval information for initially incorrect answers, filtering in beneficial positive examples and excluding detrimental negative ones. We implement a hard-negative sampling strategy to preserve the training data of the highest quality and employ RAG-based instruction fine-tuning. Experimental results across seven datasets spanning three different tasks show that our method significantly enhances the capability of LVLMs to effectively utilize multimodal retrieval information, while also improving their resilience against misleading content.

\section{Limitation}
Our method mainly has three limitations:

\begin{itemize}
    \item Our retrieval approach heavily depends on large-scale, high-quality data sources. While using only the training data as the data source is a feasible solution, the performance is slightly inferior compared to large-scale data sources in complex tasks. Future work should explore how to leverage small sample data sources for inference through retrieval.
    
    \item Despite our method having been extensively evaluated on tasks such as Visual Question Answering (VQA), image captioning, and image classification, its generalization to other visual tasks, such as image generation and image segmentation, remains unexplored. Future work should investigate the adaptability of our framework to a broader range of tasks.

    \item Given that the retrieval process currently supports a maximum of three image-caption pairs due to lengthy descriptions, future optimizations could include using shorter captions, employing methods to compress descriptions, or increasing the maximum input tokens for LVLMs. These improvements would enable more image-caption pairs to be included, enhancing the accuracy of downstream tasks of LVLMs.

\end{itemize}


\bibliography{custom}

\newpage

\appendix
\section{Appendix}
\label{sec:appendix}

\subsection{Algorithm}\label{sec:alg}

\begin{algorithm}
\caption{Robust RAG Training Framework}
\begin{algorithmic}[1]
\Require Input question $q$ and image $i_{\text{test}}$, Image-Caption collection $\mathbb{D}$, Evaluate Tools $T$, Vision Encoder $p_V^\theta$, LVLMs $M_{\theta}$, SFT data collection $C$, Training data set $S$, Positive set $\mathbb{C}_{\text{pos}}$, Negative set $\mathbb{C}_{\text{neg}}$
\State $S \gets []$
\For{each instruction $x$ in $C$}
    \State response $\gets M_{\theta}(x)$
    \State state $\gets T$(response)
    \If{not state}
        \State $S \gets S \cup \{x\}$
    \EndIf
\EndFor
\For{each instruction $x$ in $S$}
    \State $[i, c] \gets$ Retrieval from $\mathbb{D}$ with query $i_{\text{test}}$
    \State response $\gets M_{\theta}(x, [i, c])$
    \State state $\gets T$(response)
    \If{state}
        \State $C_{\text{pos}} \gets \mathbb{C}_{\text{pos}} \cup \{(x,$
        \State \hspace{1.2em} $[\max_{i_j \in \mathbb{C}_{\text{pos}}} p_V^\theta(x, i_j), c_j])\}$
    \Else
        \State $C_{\text{neg}} \gets \mathbb{C}_{\text{neg}} \cup \{(x,$
        \State \hspace{1.2em} $[\max_{i_j \in \mathbb{C}_{\text{neg}}} p_V^\theta(x, i_j), c_j])\}$
    \EndIf
\EndFor
\State $S \gets [\mathbb{C}_{\text{pos}}, \mathbb{C}_{\text{neg}}]$
\While{$M_{\theta}$ has not converged}
    \State Update parameters of $M_{\theta}$ on $S$
\EndWhile

\end{algorithmic}
\label{algorithm}
\end{algorithm}

\begin{table}[t]
\resizebox{0.49\textwidth}{!}{
\begin{tabular}{lll}
\toprule
\textbf{Dataset/Benchmark} & \textbf{Answer Type}           & \textbf{Test}  \\ \midrule
POPE              & Yes/No                & 9,000  \\
MMStar            & Multiple Choice       & 1,500  \\
Vizwiz-VQA        & Single word or Phrase & 8,000  \\
MS-COCO           & Text                  & 5,000  \\
Vizwiz-Caption    & Text                  & 7,750  \\
CIFAR-10          & Class Name            & 10,000 \\
EmoSet            & Class Name            & 800*   \\ \bottomrule
\end{tabular}
}
\caption{The statistics of the datasets used in this paper. * denotes we randomly selected 800
samples from EmoSet to constitute the test set.}
\label{tab:dataset_details}
\end{table}

\subsection{Data Analysis}
\label{sec: data_analysis}
In this section, we introduce the datasets used in our experiments. The statistics of these datasets are shown in Table \ref{tab:dataset_details}.

\paragraph{POPE}
POPE \cite{pope} offers a method to assess object hallucination in LVLMs by querying if specific objects exist in images. The queries are balanced between existent and non-existent objects (50\% each). There are three sampling settings: random, popular, and adversarial. The evaluation pivots on two key metrics: Accuracy and the F1 score.

\paragraph{MMStar}
MMStar \cite{mmstar} is an advanced benchmark designed to evaluate the capabilities of LVLMs across multiple dimensions. The benchmark includes 1,500 meticulously selected challenge samples. These samples are initially chosen from existing benchmarks using an automated pipeline, followed by a rigorous human review to ensure high quality.

\paragraph{Vizwiz-VQA}
Vizwiz-VQA \cite{vizwiz_vqa} is the task of returning the answer to a question about an image. It has 8,000 test samples with the unique label "Unanswerable."

\paragraph{MS-COCO}
The MS-COCO \cite{ms-coco} dataset is a large-scale dataset for object detection, segmentation, key-point detection, and captioning. We use this dataset only for the image captioning task.

\begin{table*}[htbp]
\resizebox{\textwidth}{!}{
\begin{tabular}{lcccccccccccc}
\toprule
            & \multicolumn{2}{c}{\textbf{POPE (R)}}   & \multicolumn{2}{c}{\textbf{POPE (P)}}   & \multicolumn{2}{c}{\textbf{POPE (A)}}   & \multicolumn{5}{c}{\textbf{MS-COCO}}                                                               & \textbf{CIFAR-10}             \\
            & Acc.          & F1            & Acc.          & F1            & Acc.          & F1            &  B@4           & METEOR        & ROUGE-L       & CIDEr          & SPICE       & Acc. \\ \midrule
Zero-shot   & 87.3          & 86.0          & 86.1          & 84.9          & 84.2          & 83.4          & 22.8                     & 28.2          & 51.4          & 85.4           & 22.2          & 81.5                 \\
            & \multicolumn{12}{c}{\cellcolor{gray!20}\textit{\textbf{1-shot}}}                                                                                                                                                                                      \\
Vanilla-RAG & 87.7          & 86.3          & 85.2          & 84.1          & 82.8          & 81.9          & 22.1                     & 27.8          & 50.7          & 76.2           & 21.9          & 79.8                 \\
\textbf{Ours}        & 89.6          & 89.1          & 87.8          & 87.4          & 83.3          & 83.7          & 26.6 & 29.4          & 54.0          & 96.2           & 23.7          & 82.4                 \\
                    & \multicolumn{12}{c}{\cellcolor{gray!20}\textit{\textbf{2-shot}}}                                                                                                                                                                                      \\
Vanilla-RAG & 87.9          & 86.5          & 86.3          & 85.0          & 83.3          & 82.5          & 24.9                     & 28.5          & 52.8          & 89.7           & 22.6          & 79.5                 \\
\textbf{Ours}        & \textbf{89.8} & \textbf{89.3} & \textbf{87.9} & \textbf{87.6} & \textbf{83.6} & \textbf{83.9} & \textbf{27.9}            & \textbf{29.9} & \textbf{55.1} & \textbf{101.3} & \textbf{24.2} & \textbf{83.5}        \\
            & \multicolumn{12}{c}{\cellcolor{gray!20}\textit{\textbf{3-shot}}}                                                                                                                                                                                      \\
Vanilla-RAG & 87.5          & 86.0          & 85.5          & 84.2          & 82.6          & 81.6          & 23.0                     & 28.1          & 51.4          & 79.2           & 22.2          & 79.1                 \\
\textbf{Ours}        & 89.3          & 88.7          & 87.8          & 87.4          & 83.2          & 83.3          & 27.1                     & 29.3          & 54.4          & 98.7           & 23.6          & 82.0                 \\ \bottomrule
\end{tabular}
}
\caption{Number of exemplars.}
\label{tab:number_of_examplrs}
\end{table*}

\begin{table*}[htbp]
\resizebox{\textwidth}{!}{
\begin{tabular}{ccccccccccccc}
\toprule
\multicolumn{1}{l}{} & \multicolumn{2}{c}{\textbf{\textbf{POPE (R)}}} & \multicolumn{2}{c}{\textbf{\textbf{POPE (P)}}} & \multicolumn{2}{c}{\textbf{\textbf{POPE (A)}}} & \multicolumn{5}{c}{\textbf{MS-COCO}}             & \textbf{CIFAR-10} \\
\multicolumn{1}{l}{} & Acc.         & F1           & Acc.         & F1           & Acc.         & F1           & B@4  & METEOR & ROUGE-L & CIDEr & SPICE & Acc.     \\ \midrule
Vanilla-RAG          & 87.9         & 86.5         & 86.3         & 85.0         & 83.3         & 82.5         & 24.5 & 28.3   & 51.4    & 79.8  & 22.4  & 79.7     \\ \midrule
1k                   & 86.6         & 84.8         & 85.8         & 84.0         & 83.2         & 81.6         & 25.3 & 26.7   & 51.0    & 98.1  & 22.7  & 79.2     \\
2k                   & \textbf{89.8}         & \textbf{89.3}         & \textbf{87.9}         & \textbf{87.6}         & \textbf{83.6}         & \textbf{83.9}         & \textbf{27.9} & \textbf{29.9}   & \textbf{55.1}    & \textbf{101.3} & \textbf{24.2}  & \textbf{83.5}     \\
3k                   & 88.8         & 87.9         & 87.3         & 86.5         & 83.2         & 82.9         & 27.5 & 29.4   & 53.5    & 99.3  & 23.9  & 81.5     \\
4k                   & 88.2         & 87.5         & 87.0         & 86.3         & 82.9         & 82.5         & 26.8 & 28.8   & 51.4    & 96.8  & 23.0  & 79.6     \\ \bottomrule
\end{tabular}
}
\caption{Effect of training data size.}
\label{tab:training_data_size}
\end{table*}

\paragraph{Vizwiz-Caption}
VizWiz-Caption \cite{vizwiz-caption} is a specialized dataset for evaluating and improving image captioning systems, particularly for visually impaired users. It consists of images taken by visually impaired individuals using their smartphones, accompanied by human-generated captions.

\paragraph{CIFAR-10}
CIFAR-10 \cite{cifar-10} is a well-known benchmark dataset primarily used for evaluating image classification algorithms. The dataset is split into 50,000 training images and 10,000 test images, divided into ten different classes: airplanes, automobiles, birds, cats, deer, dogs, frogs, horses, ships, and trucks.

\paragraph{EmoSet}
EmoSet \cite{emoset} comprises 3.3 million images in total, with 118,102 of these images carefully labeled by human annotators, making it five times larger than the largest existing dataset. We randomly sampled 100 instances from each class to serve as our test set.

\subsubsection{Metrics}
Unless otherwise specified, we use exact match as the evaluation metric for VQA and classification tasks. For captioning tasks, we use BLEU-4, ROUGE-L, CIDEr, METEOR, and SPICE as evaluation metrics\footnote{We use the official COCO evaluation toolkit.}.

\subsection{Additional Ablation Study and Experiment Analysis}

\subsubsection{Sensitivity to the Number of Examplars}
Table \ref{tab:number_of_examplrs} shows the performance of our model with different numbers of examples. Due to the long captions of ShareGPT-4V, only three examples can fit within a 4096 context window. Our method demonstrates robustness with 1, 2, and 3 examples, indicating adaptability to various numbers of examples. However, the performance peaks with 2 examples and declines with 1 and 3 examples. The decline with 1 example may be due to insufficient information, while 3 examples may introduce excessive irrelevant information.

\subsubsection{Effect of Training Data Size}

Table \ref{tab:training_data_size} shows the experiments on the amount of training data. Using only 2k data points, our model is already able to utilize RAG and achieve the best performance fully. Although the performance with 3k and 4k data points is slightly worse than with 2k, the results still surpass those of vanilla-RAG. This indicates that our framework can sufficiently leverage its capabilities using just 2k samples self-generated by the model.

\begin{table*}[htbp]
\resizebox{\textwidth}{!}{
\begin{tabular}{lcccccccccccc}
\toprule
            & \multicolumn{2}{c}{\textbf{POPE (R)}}   & \multicolumn{2}{c}{\textbf{POPE (P)}}   & \multicolumn{2}{c}{\textbf{POPE (A)}}   & \multicolumn{5}{c}{\textbf{MS-COCO}}                                                               & \textbf{CIFAR-10}             \\
            & Acc.          & F1            & Acc.          & F1            & Acc.          & F1            &  B@4           & METEOR        & ROUGE-L       & CIDEr          & SPICE       & Acc. \\ \midrule
Vanilla-RAG        &              &              &              &              &              &              &      &        &         &       &       &          \\
w/ image-caption & 87.9         & 86.5         & 86.3         & 85.0         & 83.3         & 82.5         & 24.5 & 28.3   & 51.4    & 79.8  & 22.4  & 79.7     \\
w/ caption         & 87.4         & 86.3         & 85.9         & 84.7         & 83.0         & 82.3         & 24.7 & 28.5   & 51.8    & 80.6  & 22.9  & 79.2    \\ \bottomrule
\end{tabular}
}
\caption{Performance of Vanilla-RAG on downstream tasks with different retrieval content.}
\label{tab:image_caption}
\end{table*}

\begin{table*}[htbp]
\resizebox{\textwidth}{!}{
\begin{tabular}{lccccc|ccccc|c}
\toprule
            & \multicolumn{5}{c}{\textbf{MS-COCO}}                                                       & \multicolumn{5}{c}{\textbf{Vizwiz-Caption}}  & \multirow{2}{*}{\textbf{Avg.}} \\
           & B@4           & METEOR        & ROUGE-L       & CIDEr         & SPICE           & B@4           & METEOR        & ROUGE-L       & CIDEr         & SPICE         &  \\ \midrule
\multicolumn{12}{c}{\cellcolor{gray!20}\textit{\textbf{7B Parameter Model}}} \\ 
Zero-shot   & 22.3          & 28.0          & 50.9          & 75.3          & 22.1            & 15.2          & 19.3          & 40.9          & 47.3          & 11.8          & 33.51 \\
Vanilla-RAG & 24.5          & 28.3          & 51.4          & 79.8          & 22.4            & 21.0          & 21.6          & 45.2          & 67.1          & 14.7          & 37.60 \\
Rerank-RAG  & 24.7          & 28.6          & 52.0          & 82.1          & 22.6            & 20.5          & 20.9          & 44.3          & 64.6          & 13.9          & 37.42 \\
Filter-RAG  & 26.8          & 29.4          & 54.2          & 97.0          & 23.7            & 21.4          & 21.8          & 45.9          & 68.0          & 14.9          & 40.31 \\
\textbf{Ours} & \textbf{27.9} & \textbf{29.9} & \textbf{55.1} & \textbf{101.3}& \textbf{24.2}   & \textbf{22.4} & \textbf{22.3} & \textbf{46.3} & \textbf{71.1} & \textbf{15.3} & \textbf{43.57} \\ \midrule
\multicolumn{12}{c}{\cellcolor{gray!20}\textit{\textbf{13B Parameter Model}}} \\
Zero-shot   & 22.8          & 28.2          & 51.4          & 85.4          & 22.2            & 18.0          & 19.8          & 42.5          & 57.2          & 12.7          & 36.52 \\
Vanilla-RAG & 24.9          & 28.5          & 52.8          & 89.7          & 22.6            & 21.6          & 21.3          & 45.4          & 59.4          & 13.2          & 37.94 \\
Rerank-RAG  & 25.1          & 28.6          & 53.5          & 93.1          & 22.8            & 21.8          & 21.2          & 45.5          & 60.2          & 13.4          & 38.52 \\
Filter-RAG  & 25.5          & 28.7          & 53.9          & 95.6          & 23.0            & 22.0          & \textbf{21.4} & 45.6          & 61.5          & 13.6          & 39.08 \\
\textbf{Ours} & \textbf{30.8} & \textbf{29.1} & \textbf{56.0} & \textbf{111.5}& \textbf{23.5}   & \textbf{24.5} & 21.1 & \textbf{46.2} & \textbf{71.2} & \textbf{14.5} & \textbf{44.83} \\ \bottomrule
\end{tabular}
}
\caption{Full results of 7B and 13B Robust-LlaVA on MS-COCO and Vizwiz-Caption. The best performance in the table is highlighted in \textbf{bold}.}
\label{tab:detail_results}
\end{table*}

\subsubsection{Effect of Different Retrieved Content}
In this section, we explore the performance differences when using image-caption pairs versus using only captions for retrieval across three tasks, as shown in Table \ref{tab:image_caption}. For VQA and classification tasks, using both image and caption yields the best results, as the additional information from the image is beneficial for tasks that require a strong understanding of the image. However, for the captioning task, using only captions performs better since this task requires the model to generate a relevant response based solely on the retrieved descriptions.

\subsubsection{Detail Results of Captioning Tasks}
In the main table, the metrics for the captioning task are the sum of BLEU-4, METEOR, ROUGE-L, CIDEr, and SPICE. We present the detailed results of our 7B and 13B models on MS-COCO and Vizwiz-Caption in Table \ref{tab:detail_results}.

\subsubsection{Effect of Irrelevant Content}
\begin{figure*}[htbp]
    \centering
    \subfigure[Performance of the base model (Qwen-VL) without using RAG (Base), RAG with irrelevant content (Irrelevant), and RAG on POPE-popular, MS-COCO, and CIFAR-10.]{
        \includegraphics[width=0.45\textwidth]{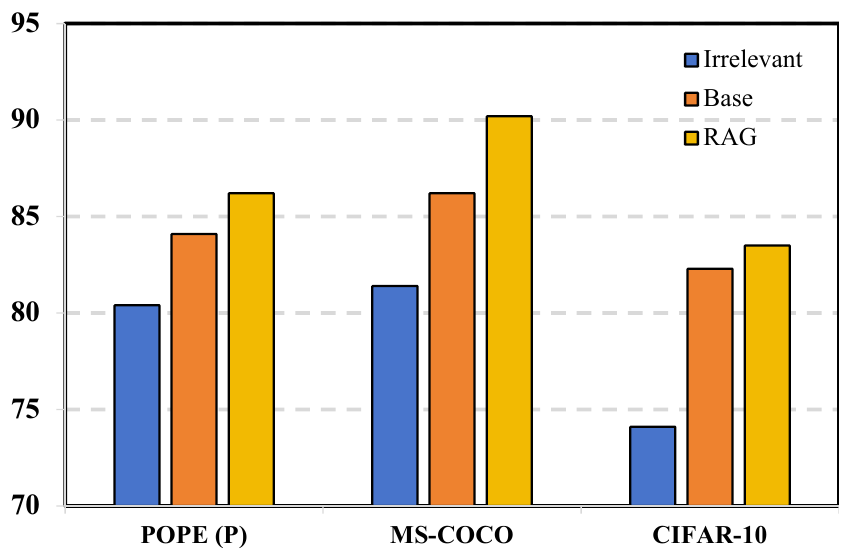}
        \label{fig:qwen}
    }
    \hfill
    \subfigure[Performance of the base model (mPLUG-Owl2) without using RAG (Base), RAG with irrelevant content (Irrelevant), and RAG on POPE-popular, MS-COCO, and CIFAR-10.]{
        \includegraphics[width=0.45\textwidth]{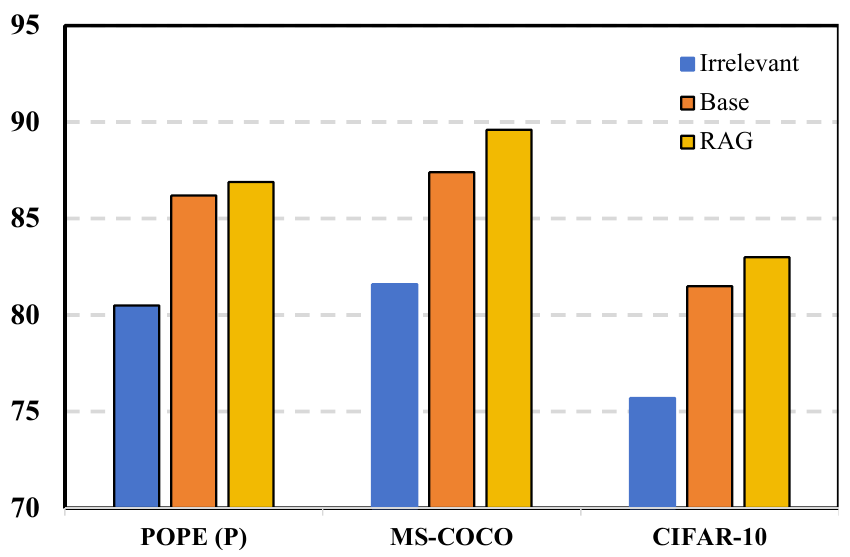}
        \label{fig:mplug}
    }
    \caption{Ablation Study of Database Size and Data Filter.}
    \label{fig:irrelevant_content}
\end{figure*}

We also tested Qwen-VL \cite{qwen-vl} and mPLUG-Owl2 \cite{mplug-owl} under three settings (Base, Irrelevant, and RAG) across three tasks. The results are shown in Figures \ref{fig:qwen} and \ref{fig:mplug}. It can be observed that irrelevant content has a significant impact on the current LVLMs.

\subsection{Case Study}

\begin{figure*}[htbp]
\centering
\includegraphics[width=0.8\linewidth]{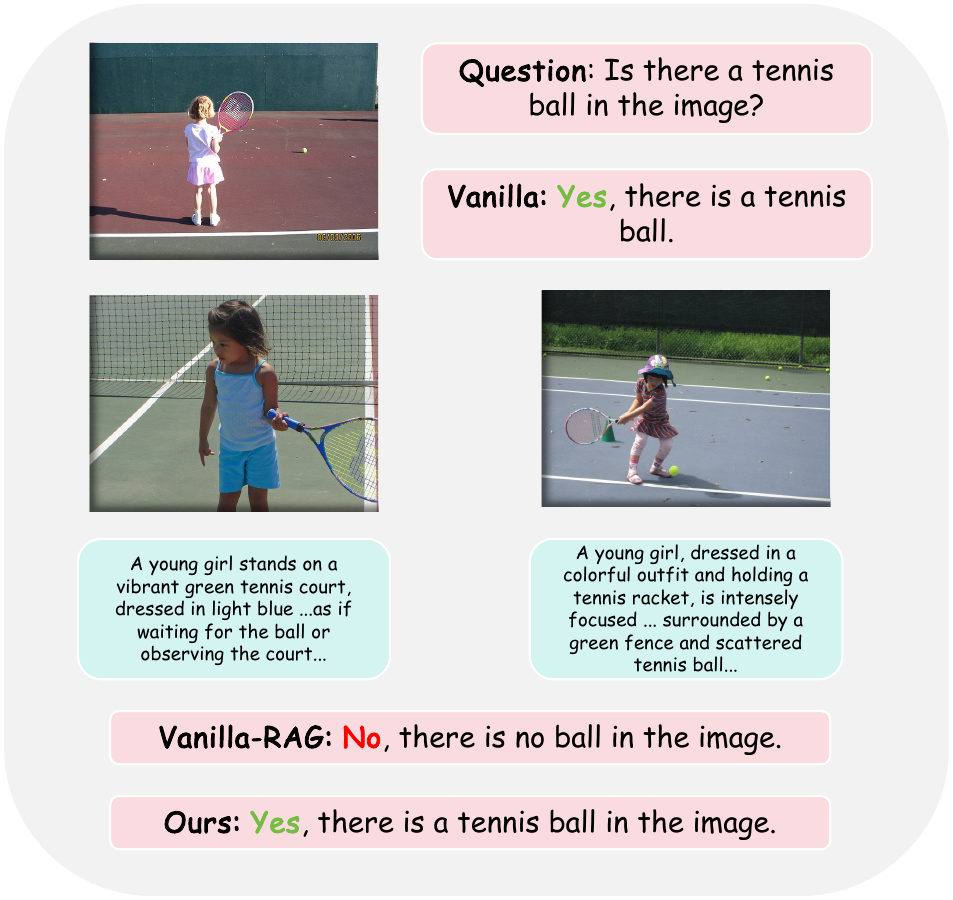}
\caption{Case for comparing our method with zero-shot and vanilla-RAG.}
\label{fig:case1}
\end{figure*}

\begin{figure*}[htbp]
\centering
\includegraphics[width=0.8\linewidth]{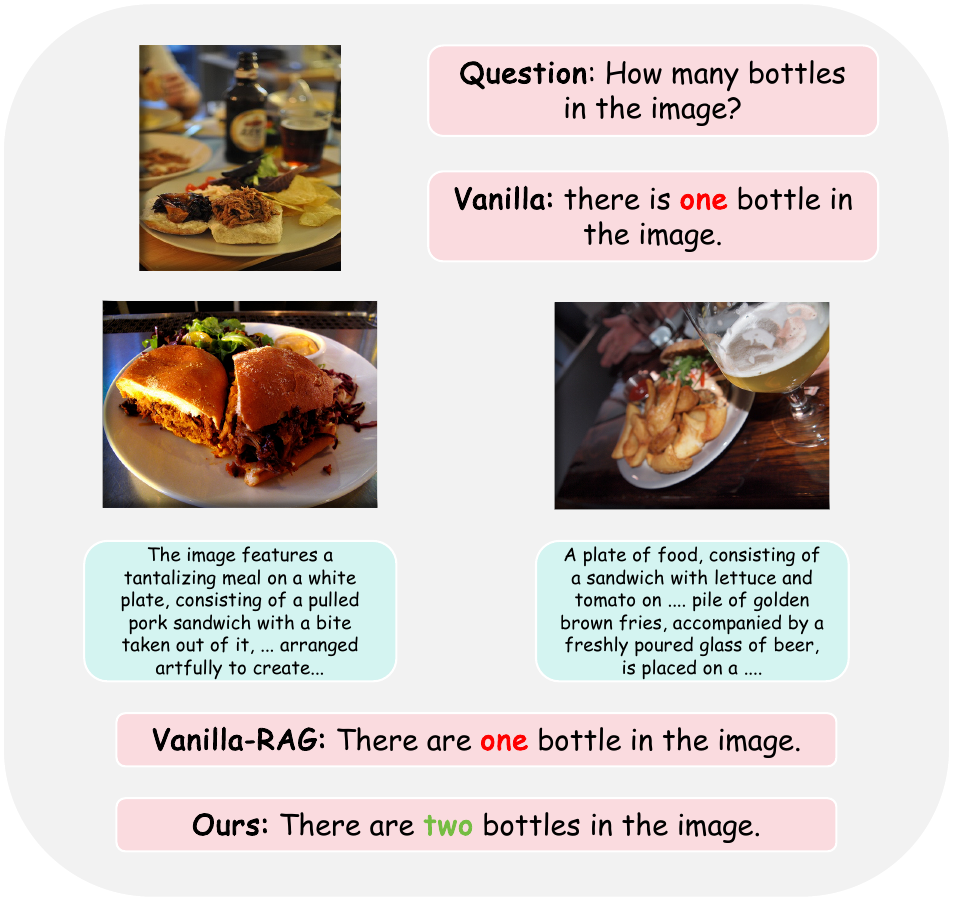}
\caption{Case for comparing our method with zero-shot and vanilla-RAG.}
\label{fig:case2}
\end{figure*}

\begin{figure*}[htbp]
\centering
\includegraphics[width=0.8\linewidth]{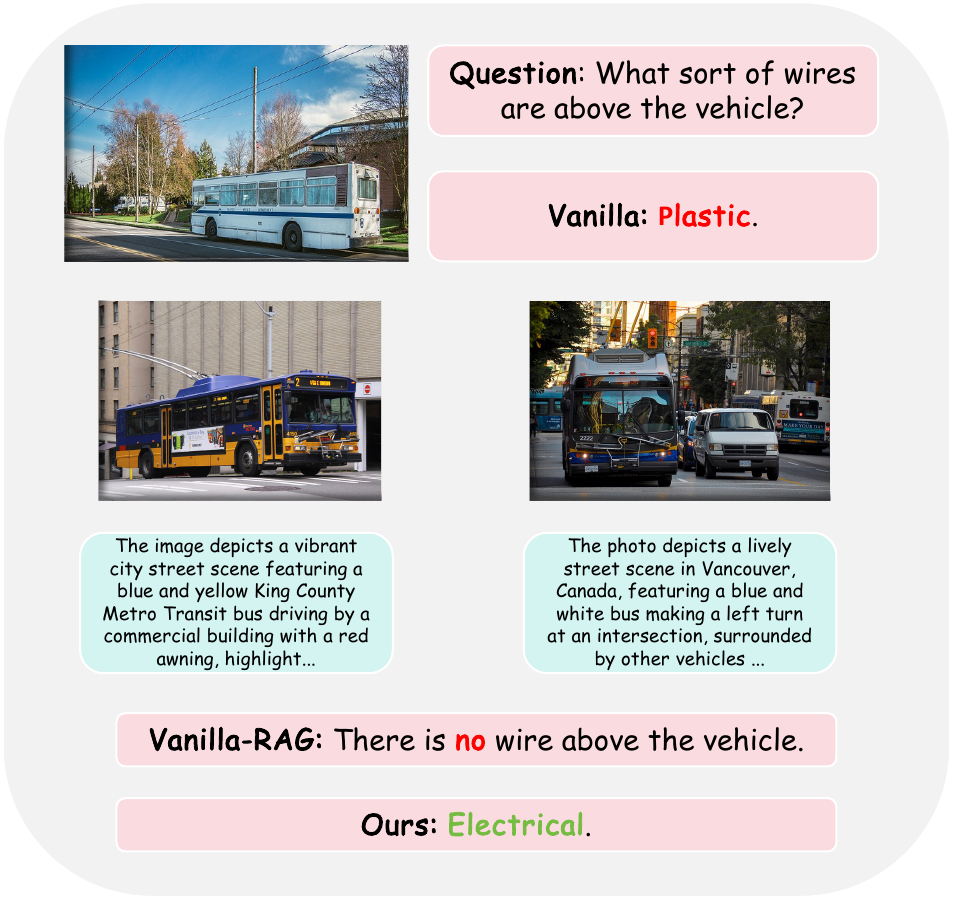}
\caption{Case for comparing our method with zero-shot and vanilla-RAG.}
\label{fig:case3}
\end{figure*}

\begin{figure*}[htbp]
\centering
\includegraphics[width=0.8\linewidth]{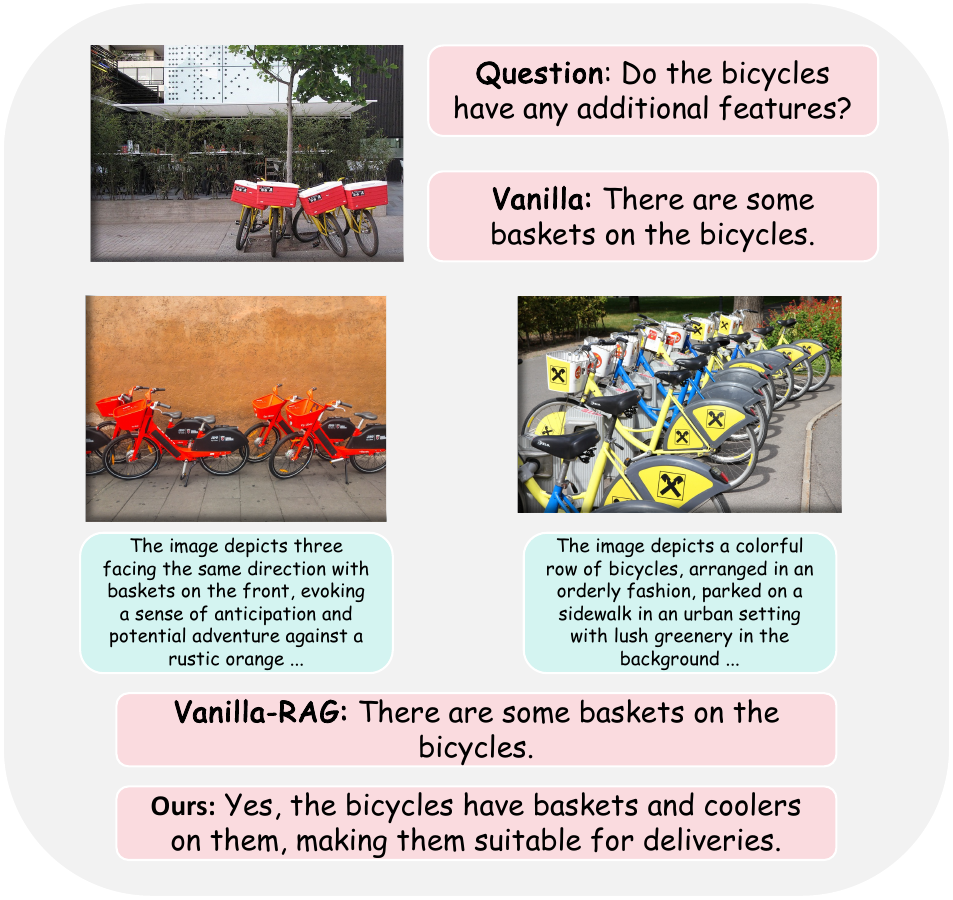}
\caption{Case for comparing our method with zero-shot and vanilla-RAG.}
\label{fig:case4}
\end{figure*}

We present four examples comparing our method with zero-shot and vanilla-RAG, as shown in Figures \ref{fig:case1}, \ref{fig:case2}, \ref{fig:case3}, and \ref{fig:case4}.

\end{document}